\documentclass[11pt]{article}
\usepackage{amssymb} 
\usepackage[preprint]{acl}
\usepackage{times}
\usepackage{latexsym}

\usepackage[T1]{fontenc}

\usepackage[utf8]{inputenc}

\usepackage{microtype}

\usepackage[utf8]{inputenc}
\usepackage{booktabs}
\usepackage[table]{xcolor}
\usepackage{geometry}
\usepackage[symbol]{footmisc}
\usepackage{inconsolata}

\usepackage{graphicx}
\usepackage{float}
\usepackage{amsmath}
\usepackage{algorithm}
\usepackage{algpseudocode}
\usepackage{multirow}
\usepackage{tcolorbox}
\usepackage{colortbl}
\definecolor{forward}{RGB}{84, 130, 53}
\definecolor{inverse}{RGB}{47, 85, 151}
\definecolor{resist}{RGB}{128, 0, 128}
\definecolor{rebound}{RGB}{133, 19, 33}

\definecolor{def}{RGB}{119, 228, 200}
\definecolor{thm}{RGB}{69, 53, 193}
\newtcolorbox{thmbox}[1][]{colback=thm!5!white,colframe=thm!60!black,boxsep=-4pt,grow to left by=4pt,left=10pt,grow to right by=4pt,right=10pt,top=10pt,bottom=10pt,#1}
\newtcolorbox{defbox}[1][]{colback=def!5!white,colframe=def!60!black,boxsep=-4pt,grow to left by=4pt,left=10pt,grow to right by=4pt,right=10pt,top=10pt,bottom=10pt,#1}
\newtheorem{theorem}{Property}
\usepackage{eso-pic}
\usepackage{graphicx}

\title{GUI-CIDER: Mid-training GUI Agents via Causal Internalization \\and Density-aware Exemplar Reselection}

\author{
    Zheng Wu\textsuperscript{1,2}\thanks{Work completed while Zheng Wu, Zhengxi Lu, Tianjie Ju, and Yanyu Chen were interns at Meituan.} \quad 
    Chengcheng Han\textsuperscript{2} \quad 
    Zhengxi Lu\textsuperscript{2,3} \quad 
    Tianjie Ju\textsuperscript{1,2} \quad 
    Yanyu Chen\textsuperscript{2,4} \\ 
    \textbf{Qi Gu}\textsuperscript{2}\thanks{Corresponding authors.} \quad 
    \textbf{Xunliang Cai}\textsuperscript{2} \quad 
    \textbf{Zhuosheng Zhang}\textsuperscript{1}\textsuperscript{\textdagger}
    \\
    \textsuperscript{1}School of Computer Science, Shanghai Jiao Tong University \quad 
    \textsuperscript{2}Meituan \\ 
    \textsuperscript{3}Zhejiang University \quad 
    \textsuperscript{4}The Chinese University of Hong Kong\\
    \texttt{\{wzh815918208,zhangzs\}@sjtu.edu.cn} \quad 
    \texttt{guqi03@meituan.com}
}

\begin{document}

\AddToShipoutPictureBG*{
  \AtPageUpperLeft{
    \put(\LenToUnit{2.7cm},\LenToUnit{-2.5cm}){
      \includegraphics[width=3cm]{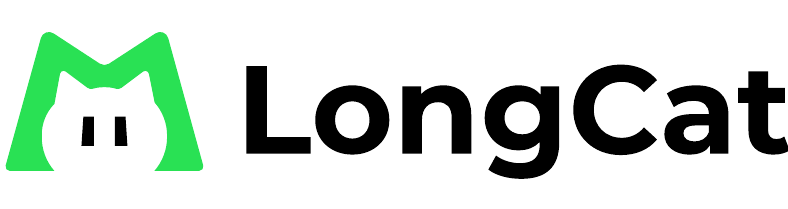}
    }
  }
}
\maketitle


\begin{abstract}
Despite the rapid progress of multimodal large language models in building Graphical User Interface (GUI) agents, their real-world task completion is fundamentally bottlenecked by a lack of world knowledge about GUI operations. 
Existing solutions typically rely on expensive multi-agent scaffolding or conventional post-training paradigms, such as Supervised Fine-Tuning (SFT) and Reinforcement Learning (RL). 
However, post-training only allows agents to \textbf{implicitly} absorb world knowledge through action annotations or reward signals, leading to inefficient trajectory memorization rather than genuine comprehension. 
Therefore, an approach that enables \textbf{explicit} learning of this knowledge is imperative. 
To this end, we propose \textbf{GUI-CIDER}, a mid-training method that explicitly internalizes GUI world knowledge through \textbf{C}ausal \textbf{I}nternalization and \textbf{D}ensity-aware \textbf{E}xemplar \textbf{R}eselection. 
\textsc{GUI-CIDER} operates in three stages: (1) data synthesis, which distills static planning and dynamic causal knowledge from GUI trajectories into text; (2) exemplar reselection, which filters the corpus by rewarding causal structures and penalizing semantic redundancy; and (3) mid-training, where the refined data is used to embed the acquired knowledge. 
Extensive experiments on two GUI knowledge benchmarks and three task completion benchmarks demonstrate that \textbf{GUI-CIDER} consistently improves both the agent's understanding of GUI operations and its task success rates.The codes are available at \url{https://github.com/Wuzheng02/GUI-CIDER}.
\end{abstract}

\section{Introduction}
With the rapid advances of multimodal large language models (MLLMs) in reasoning~\cite{Qwen3-VL}, planning~\cite{wei2025plangenllms,chen2026trace}, perception~\cite{yu2025auto}, and decision-making~\cite{sun2025llm}, MLLM-based Graphical User Interface (GUI) agents~\cite{tang2025survey} can now follow user instructions to autonomously control digital devices (e.g., computers~\cite{sager2026comprehensive} and smartphones~\cite{wu2025quick}) by simulating human actions (e.g., clicking and scrolling).

Existing work on GUI agents improves element grounding~\cite{liu2026infigui,tang2026gui} and task completion~\cite{bai2024digirl, xu2025mobilerl} through post-training methods such as supervised fine-tuning (SFT)~\cite{zhang2024you,ma2024coco} and reinforcement learning (RL)~\cite{lu2026ui,luo2025gui}.
However, studies~\cite{shi2025gui,li2025using} point out that as GUI agents continue to advance, the real capability bottleneck increasingly stems from \textbf{a lack of world knowledge related to GUI operations}.

\begin{figure}[t]
    \centering
    \includegraphics[width=1\linewidth]{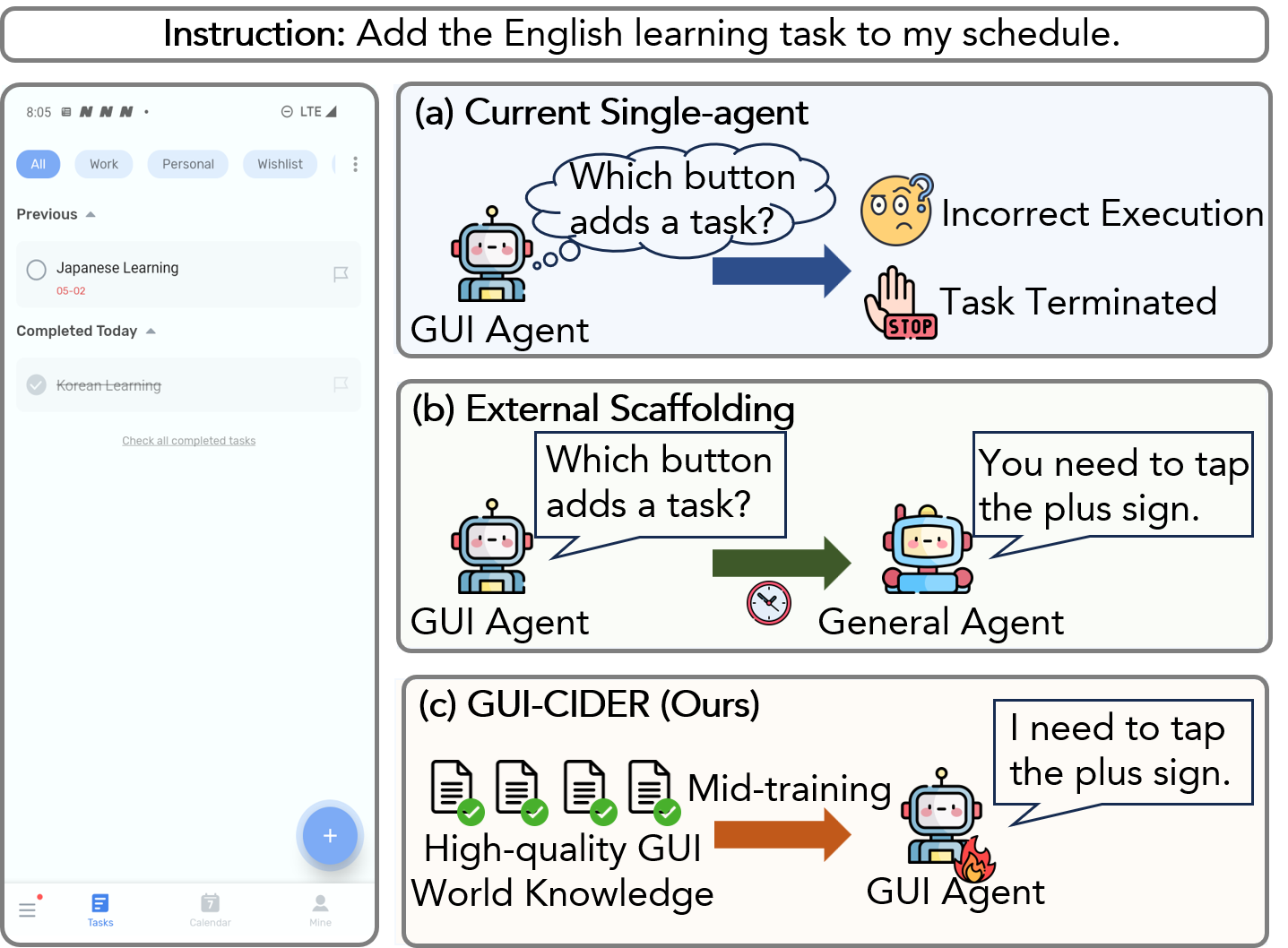}
     \caption{The motivation of GUI-CIDER. (a) Current single-agent methods lack GUI world knowledge and cannot understand that the plus sign represents adding a task. (b) External scaffolding can obtain low-level instructions by calling a general agent, but this is time-consuming. (c) GUI-CIDER enables a single agent to accomplish the task through mid-training on high-quality GUI world knowledge.}
    \vspace{-0.3cm}
    \label{fig:teaser}
\end{figure}

Although plugging a capable general-purpose model into a multi-agent system~\cite{yang2025gta1,wang2024mobile} can compensate for GUI agents' deficiency in world knowledge, it introduces additional overhead and scaffolding. 


In contrast, internalizing world knowledge within the agent is more efficient, yet conventional post-training (SFT/RL) only \textbf{implicitly} encodes such knowledge through action labels or reward signals, encouraging trajectory memorization rather than genuine comprehension.

An approach that enables \textbf{explicit} learning is therefore imperative.
  
Consequently, as shown in Figure~\ref{fig:teaser}, we propose GUI-CIDER, a mid-training method for GUI agents that explicitly internalizes world knowledge into them through \textbf{C}ausal \textbf{I}nternalization and \textbf{D}ensity-aware \textbf{E}xemplar \textbf{R}eselection.

GUI-CIDER consists of three stages: (1) data synthesis stage, (2) exemplar reselection stage, and (3) mid-training stage. 
In the \textbf{data synthesis} stage, GUI-CIDER employs a dedicated synthesis pipeline to generate static planning knowledge and dynamic causal knowledge for the GUI agent domain from publicly available GUI agent datasets~\cite{li2024effects,lu2025guiodyssey,zhang2024android}.
In the \textbf{exemplar reselection} stage, GUI-CIDER filters the data produced in the previous stage through causal-informed retention and relative density estimation based on 
$K$-nearest neighbors, resulting in a high-quality corpus that exhibits strong reasoning structures and low redundancy.
In the \textbf{mid-training} stage, GUI-CIDER uses this high-quality corpus to train the GUI agent via mid-training, thereby explicitly internalizing world knowledge into the GUI agent.

We conduct extensive experiments on three benchmarks~\cite{li2024effects,lu2025guiodyssey,zhang2024android} for GUI agent task completion and two benchmarks~\cite{wang2025mmbench, shi2025gui} for GUI agent knowledge.
Experimental results show that GUI-CIDER achieves an average relative improvement of 9.70\% in task success rate compared to post-training baselines. 
Meanwhile, on the GUI knowledge bench, it enables an 8B-scale agent to reach a level close to that of Claude-Sonnet-4.5.

Additionally, through model comparison analysis, we show that the target of mid-training should be general agents rather than one that has been excessively post-trained specifically in the GUI agent domain.
Furthermore, we validate the rationality of the GUI-CIDER pipeline through ablation studies.

To summarize, our contributions are three-fold:


(i) We propose GUI-CIDER, a mid-training method for GUI agents that explicitly internalizes world knowledge relevant to GUI agents into them through \textbf{C}ausal \textbf{I}nternalization and \textbf{D}ensity-aware \textbf{E}xemplar \textbf{R}eselection.

(ii) We contribute a corpus of approximately 100M tokens generated from the data synthesis stage of GUI-CIDER, offering a valuable resource for related research in the community.

(iii) Through extensive experiments, we demonstrate that GUI-CIDER can not only improve GUI agents' world knowledge of GUI operations but also enhance their task completion performance.

\begin{figure*}[ht]
    \centering
\includegraphics[width=1\linewidth]{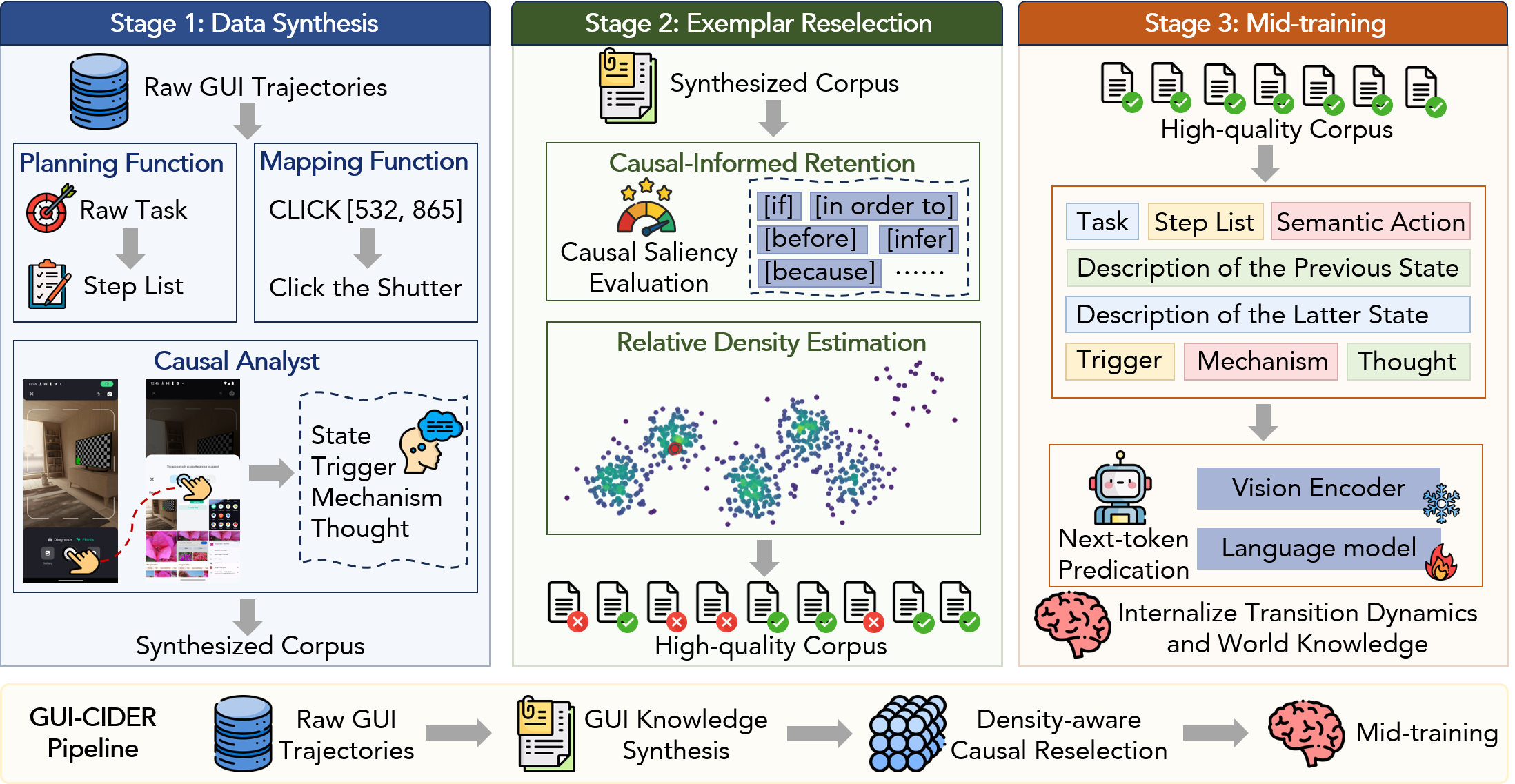}
    \caption{The pipeline of GUI-CIDER. In the data synthesis stage, GUI-CIDER synthesizes GUI world knowledge from raw GUI trajectories. In the exemplar reselection stage, GUI-CIDER selects a high-quality corpus with strong reasoning structures and low redundancy from the synthesized corpus. In the mid-training stage, GUI-CIDER internalizes the GUI world knowledge into the GUI agent through mid-training on the high-quality corpus.}
    \label{fig:pipeline}
\end{figure*}

\section{Related Work}
In this section, we first introduce recent improvements in GUI agents, and then we introduce related work on mid-training of (M)LLMs.

\subsection{GUI Agents}
GUI agents are a type of agent that operate intelligent terminals such as computers~\cite{sager2026comprehensive}, web~\cite{he2024webvoyager}, and smartphones~\cite{zhang2024you,wu2025quick} by simulating human actions like clicking and scrolling~\cite{tang2025survey, hu2025agents}.
Existing work can be broadly divided into two categories for constructing GUI agents: single-agent based and multi-agent system based.
Single-agent based GUI agents are typically developed through pre-training and post-training. Pretraining enhances the agent's perception~\cite{ma2024coco} and grounding capabilities~\cite{wu2025atlas}.
Post-training methods, on the other hand, primarily improve the agent's task completion ability through techniques such as SFT~\cite{wu2025see} and RL~\cite{lu2026ui, zhou2026gui,tang2026gui}.
Multi-agent system based GUI agents distribute capabilities such as planning~\cite{wang2024mobile}, reflection~\cite{li2026mobileuse}, and execution~\cite{yang2025gta1,agasheagent} across different agents to adapt to different tasks.
However, few existing works enhance the world knowledge of GUI agents through mid-training.

\subsection{Mid-training for (M)LLM}
Mid-training serves as a bridge~\cite{tu2025survey,mo2025mid} between pre-training and post-training, extending knowledge into specialized domains while preserving the general capabilities acquired during pre-training.
Existing (M)LLMs~\cite{team2025longcat,hu2024minicpm,liu2024deepseek} conduct data collection, data synthesis, data selection, and data decontamination from high-quality mathematical~\cite{paster2024openwebmath,han2024infimm}, QA~\cite{weimagicoder,ding2023enhancing} and coding~\cite{kocetkovstack,lozhkov2024starcoder,luo2024wizardcoder} domains.
However, there is still very little work on internalizing domain knowledge for GUI agents through mid-training.
UI-Venus-1.5~\cite{team2026ui} employed mid-training but did not open-source the data or provide specific details.
Therefore, it is valuable to explore how GUI agents can internalize knowledge through mid-training.

\section{GUI-CIDER}
In this section, we introduce GUI-CIDER, a mid-training method for GUI agents, which stands for \textbf{C}ausal \textbf{I}nternalization and \textbf{D}ensity-aware \textbf{E}xemplar \textbf{R}eselection. 
As shown in Figure~\ref{fig:pipeline}, GUI-CIDER consists of three stages: data synthesis, exemplar reselection, and mid-training. 
Next, we will introduce each of these stages in order.

\subsection{Stage 1: Data Synthesis}
Given a raw GUI agent domain dataset $\mathcal{D} = \{ \tau_1, \tau_2, \dots, \tau_N \}$, where each trajectory $\tau$ consists of a task instruction $T$ and a sequence of screenshots and actions $\{(s_i, a_i)\}_{i=1}^L$, we synthesize an augmented, knowledge-rich sample $x$. 
Specifically, the synthesized sample $x$ encompasses two primary dimensions: static planning knowledge and dynamic causal knowledge.

\paragraph{Static Planning Knowledge Extraction.}
To operationalize hierarchical task decomposition, we leverage a high-capacity LLM as a latent knowledge prior, formalizing the planning process as a structured reasoning task. 
Specifically, the planning function $\mathcal{P}(\cdot)$ is instantiated by an expert model that performs zero-shot reasoning to generate a hierarchical decomposition:
\begin{equation}
    S = \mathcal{P}(T; \mathcal{M}_{exp}) = \{g_1, g_2, \dots, g_n\},
\end{equation}
where $\mathcal{M}_{exp}$ denotes the expert reasoning engine and $g_j$ represents a high-level sub-goal in natural language. This transformation converts abstract user intent into an actionable execution graph, providing dense supervisory signals for the agent's long-term planning.

\paragraph{Dynamic Causal Knowledge Synthesis.}
To explicitly model environment transition dynamics and decision-making logic while producing a purely textual knowledge sample, we reformulate knowledge extraction as a text-grounded semantic and causal induction process. This is achieved through two specialized reasoning modules:

(i) Semantic Behavioral Grounding: A mapping function $\mathcal{B}(a_t, v_t)$ that translates raw, low-level action primitives $a_t$ and their corresponding UI metadata $v_t$ (e.g., view hierarchy) into human-interpretable semantic descriptions $a^{nl}_t$. This stage bridges the gap between discrete pixel-level coordinates and high-level functional intent.

(ii) Textual State Abstraction and Causal Logic Induction: The visual screenshots $s_{t-1}$ and $s_t$ are first converted into natural language state descriptions $d_{t-1}$ and $d_t$ through a vision-language interface. For each transition under task $T$, we then employ a causal analyst $\mathcal{C}(\cdot)$ that operates solely over textual representations. By prompting the expert model to perform retrospective and counterfactual analysis on the described states, we extract the underlying transition logic in a self-contained textual rationale $R_t$:
\begin{equation}
\begin{split}
    R_t &= \mathcal{C}(T, d_{t-1}, a^{nl}_t, d_t \mid \mathcal{M}_{exp}) \\
        &= \{ d_{t-1}, d_t, Trig_t, Mech_t, CoT_t \},
\end{split}
\end{equation}
where $Trig_t$, $Mech_t$, and $CoT_t$ denote the action trigger, the underlying UI mechanism, and the chain-of-thought rationale, respectively. The state descriptions $d_{t-1}$ and $d_t$ are explicitly stored as part of $R_t$, making the rationale self-contained and eliminating the need for raw screenshots in the final sample.

The final synthesized sample $x$ is thus defined as a purely textual, knowledge-rich tuple:
$x = \langle T, S, a^{nl}_t, R_t \rangle$.

\subsection{Stage 2: Exemplar Reselection}

To refine the synthesized corpus $\mathcal{X} = \{x_1, \dots, x_M\}$, we employ density-aware exemplar reselection. Let $\phi(x)$ be the embedding of sample $x$ in a latent space $\mathcal{Z} \subseteq \mathbb{R}^d$.

\paragraph{Causal-Informed Retention.}
Following existing work~\cite{chen2026molecular}, we first define a causal saliency function $f(x)$ based on the count of causal-logic tokens:
\begin{equation}
    f(x) = \tanh\left(\frac{\mathcal{K}(x)}{\gamma}\right),
\end{equation}
where $\mathcal{K}(x)$ denotes the count of causal-logic tokens in $R_t$ and $\gamma$ controls the causal scaling.
Here, causal-logic tokens broadly encompass words and phrases carrying causal or logical semantics (e.g., 'if', 'unless', 'because', 'due to').
Detailed causal-logic keywords can be found in Appendix~\ref{sec:appendix_artifacts}.

\paragraph{Relative Density Estimation.}
The local density $d(x)$ is defined based on the ratio of the $K$-nearest neighbor distance to the global mean distance. 
Let the raw ratio be
\begin{equation}
    r(x) = \frac{\frac{1}{K} \sum_{z \in \text{KNN}(\phi(x))} \|\phi(x) - z\|^2}
                {\frac{1}{M} \sum_{z' \in \mathcal{X}} \|\phi(x) - z'\|^2}.
\end{equation}
To obtain a density score in $[0,1]$, we apply min-max normalization across all samples in the feature set $\mathcal{X}$:
\begin{equation}
    d(x) = \frac{r(x) - \min_{x' \in \mathcal{X}} r(x')}
                 {\max_{x' \in \mathcal{X}} r(x') - \min_{x' \in \mathcal{X}} r(x')}.
\end{equation}

The retention probability $g(x)$ for each sample $x$ is then given by a non-linear combination of its semantic density $d(x)$ and the causal saliency $f(x)$:
\begin{equation}
    g(x) = \frac{1}{1 + \alpha d(x)} + \lambda \cdot f(x) \cdot \left( 1 - \frac{1}{1 + \alpha d(x)} \right),
\end{equation}
where $\alpha$ is a hyperparameter governing density sensitivity, and $\lambda \in [0, 1]$ is the weight for causal importance.

Finally, the high-quality corpus $\mathcal{X}_{\text{high}}$ is formed by retaining each sample $x$ with probability $g(x)$:
\begin{equation}
    \mathcal{X}_{\text{high}} = \{\, x \in \mathcal{X} \mid \xi_x \leq g(x) \,\},
\end{equation}
where $\xi_x \sim \text{Uniform}(0,1)$ is sampled independently for every $x$.

\subsection{Stage 3: Mid-training}
In the mid-training stage, we directly perform next-token prediction on the high-quality corpus $\mathcal{X}_{\text{high}}$.
For each synthesized sample $x$, we first format it into a single token sequence by concatenating its components in a fixed order. No distinction is made between input and output: the entire sequence is treated as a plain text stream for autoregressive language modeling.
The training objective is the standard causal language modeling loss over all tokens in the sequence:
\begin{equation}
    \mathcal{L}_{\text{mid}} = - \sum_{x \in \mathcal{X}_{\text{high}}}
    \sum_{i=1}^{L_x} \log P_{\theta}(y_i \mid y_{<i}),
\end{equation}
where $L_x$ is the total number of tokens in the serialized sequence of sample $x$, and $y_i$ denotes the $i$-th token.
By optimizing $\mathcal{L}_{\text{mid}}$, the model internalizes the transition dynamics $P(s_t \mid s_{t-1}, a_t)$ and the underlying world knowledge directly into its parametric memory, achieving causal internalization without necessitating external runtime scaffolds.

\section{Is the Retention Function \(g(x)\) a Good Function?}
In this section, we first introduce the properties of a good retention function \(g(x)\) under our task setting, and then provide theoretical support to prove that GUI-CIDER's retention function \(g(x)\) satisfies all these properties.

\subsection{Properties for the Retention Function \(g(x)\)}
To effectively select high-value samples with strong reasoning structures and low redundancy, the retention function \(g(x)\) should possess the following four properties:

\begin{thmbox}
\begin{theorem}[Causal Monotonicity]
\(g(x)\) should be monotonically increasing with respect to the causal saliency \(f(x)\). 
\end{theorem}
\end{thmbox}
Samples with more causal-logic tokens contain richer reasoning structures and therefore deserve higher retention probabilities.


\begin{thmbox}
\begin{theorem}[Density Penalty]
\(g(x)\) should be monotonically decreasing with respect to the density \(d(x)\).
\end{theorem}
\end{thmbox}
Higher density indicates that many different samples share similar semantics, leading to redundancy; thus, the retention probability should be penalized accordingly.

\begin{thmbox}
\begin{theorem}[Density Order Preservation]
\(g(x)\) should satisfy \(\frac{\partial}{\partial d}\big(d \cdot g(x)\big) > 0\).
\end{theorem}
\end{thmbox}
Although we filter the corpus, we must not invert the original density ordering of the semantic space, thereby preserving the relative density structure.


\begin{thmbox}
\begin{theorem}[Density-Causal Synergy]
$g(x)$ should satisfy $\frac{\partial^2 g(x)}{\partial f \partial d} > 0$.
\end{theorem}
\end{thmbox}
In denser regions where semantic redundancy is high, the survival competition is fiercer. 
Therefore, an increase in causal saliency should provide a greater marginal benefit to the retention probability, enabling the most logically rigorous exemplars to stand out among highly redundant samples.

\subsection{Theoretical Support}
We now prove that the retention function defined in GUI-CIDER satisfies the three properties.

\paragraph{Proof of Property 1.}
For a fixed density \(d\), the derivative of \(g\) with respect to \(f\) is
\begin{equation}
    \frac{\partial g}{\partial f} = \lambda \left( 1 - \frac{1}{1 + \alpha d} \right) = \lambda \frac{\alpha d}{1 + \alpha d}.
\end{equation}

Since \(\lambda \ge 0\), \(\alpha > 0\), and \(d \ge 0\), we have \(\frac{\partial g}{\partial f} \ge 0\). Thus, \(g(x)\) is monotonically non-decreasing in \(f(x)\). For any sample with \(d > 0\), the derivative is strictly positive, ensuring that higher causal saliency strictly increases the retention probability.

\paragraph{Proof of Property 2.}
The derivative with respect to density \(d\) is
\begin{align}
    \frac{\partial g}{\partial d} = -\frac{\alpha(1 - \lambda f)}{(1 + \alpha d)^2}.
\end{align}

Because \(f \in [0,1)\) and \(\lambda \in [0,1]\), we have \(1 - \lambda f \ge 0\). With \(\alpha > 0\), the derivative is non-positive, so \(g(x)\) is monotonically non-increasing in \(d(x)\). This directly imposes a redundancy penalty: denser samples receive lower retention probabilities.

\begin{table*}[t]
\centering
\begin{tabular}{llccc}
\toprule
\textbf{Task Category} & \textbf{Dataset} & \textbf{Evaluation Format} & \textbf{Train Size} & \textbf{Test Size} \\
\midrule
\multirow{2}{*}{GUI Agent Knowledge} & MMBench-GUI L1 & MCQ & - & 3,561 \\
 & GUI Knowledge Bench & T/F, MCQ & - & 3,483 \\
\midrule
\multirow{3}{*}{GUI Agent Task Completion} & AITZ & \multirow{3}{*}{Action Generation} & 13,919 & 4,723 \\
 & AndroidControl & & 69,670 & 7836 \\
 & GUI-Odyssey & & 102,086 & 25,807 \\
\bottomrule
\end{tabular}
\caption{Overview of datasets and evaluation format.}
\label{tab:dataset_overview}
\end{table*}

\begin{table*}[t]
\centering
\setlength{\tabcolsep}{6pt}
\begin{tabular}{lccccccccc}
\toprule
& \multicolumn{3}{c}{AITZ} 
& \multicolumn{3}{c}{AndroidControl}
& \multicolumn{3}{c}{GUI-Odyssey} \\
\cmidrule(lr){2-4}
\cmidrule(lr){5-7}
\cmidrule(lr){8-10}
\textbf{Method} & Type & SR & TSR
& Type & SR & TSR
& Type & SR & TSR \\
\midrule
\rowcolor{gray!15} \multicolumn{10}{c}{\textbf{Qwen3-VL-4B-Instruct}} \\
Zero-shot      & 59.58 & 39.53 & 0.79 & 74.92 & 51.82 & 13.60 & 61.89 & 41.00 & 0.30  \\
GUI-CIDER     & 60.46 & 41.22 & 0.99 & 76.24 & 53.58 & 14.43 & 64.83 & 43.45 & 0.42 \\
Post-training            & 76.63 & 60.43 & 4.94 & 85.07 & 68.75 & 28.43 & 89.72 & 73.46 & 3.86 \\
GUI-CIDER + Post-training   & 77.43 & 61.87 & 5.14 & 85.17 & 69.77 & 28.56 & 89.47 & 75.36 & 4.34 \\
\rowcolor{gray!15} \multicolumn{10}{c}{\textbf{Qwen3-VL-8B-Instruct}} \\
Zero-shot      & 56.91 & 40.82 & 0.99 & 73.92 & 52.49 & 13.96 & 67.86 & 44.16 & 0.36 \\
GUI-CIDER      & 60.70 & 42.07 & 1.58 & 74.87 & 54.09 & 15.13 & 70.26 & 48.55 & 0.36 \\
Post-training            & 72.98 & 58.16 & 4.15 & 83.50 & 65.34 & 23.46 & 88.82 & 71.74 & 3.32 \\
GUI-CIDER + Post-training   & 73.70 & 60.33 & 5.14 & 83.41 & 66.82  & 25.35 & 89.65 & 73.36 & 3.63\\
\bottomrule
\end{tabular}
\caption{Results of AITZ, AndroidControl and GUI-Odyssey benchmarks. Whether compared with zero-shot or post-training methods, adding a mid-training process leads to improvements.}
\label{tab:threebench}
\end{table*}

\begin{table*}[t]
\centering
\setlength{\tabcolsep}{7pt}
\begin{tabular}{lccccccc}
\toprule
\textbf{Model} & Windows & MacOS & Linux & iOS & Android & Web & Overall \\
\midrule
\rowcolor{gray!15} \multicolumn{8}{c}{\textbf{Easy Level}} \\
GPT-4o & 62.47 & 67.89 & 62.38 & 58.52 & 56.41 & 58.51 & 60.16 \\
Qwen-Max-VL & 69.05 & 72.51 & 69.91 & 70.82 & 63.09 & 69.46 & 68.15 \\
Qwen2.5-VL-72B & 65.86 & 75.23 & 73.02 & 67.24 & 58.09 & 72.08 & 66.98 \\
UI-TARS-72B-DPO  & 41.59 & 28.52 & 35.16 & 31.08 & 52.25 & 35.33 & 40.18 \\
InternVL3-72B & 74.67 & 78.72 & 79.16 & 83.57 & 80.10 & 81.18 & 79.15 \\
\midrule
GUI-CIDER-8B  & 95.19 & 97.62 & 96.91 & 90.43 & 93.44 & 94.98 & 94.69  \\

\midrule
\rowcolor{gray!15} \multicolumn{8}{c}{\textbf{Medium Level}} \\
GPT-4o & 56.33 & 63.13 & 59.70 & 54.06 & 57.69 & 54.98 & 57.24 \\
Qwen-Max-VL & 63.40 & 73.85 & 66.90 & 68.02 & 63.66 & 64.59 & 65.44 \\
Qwen2.5-VL-72B & 66.29 & 72.73 & 72.63 & 59.27 & 66.24 & 68.24 & 67.45 \\
UI-TARS-72B-DPO & 38.83 & 41.60 & 37.14 & 41.72 & 54.74 & 31.55 & 41.77 \\
InternVL3-72B & 71.46 & 78.58 & 79.88 & 78.43 & 81.36 & 78.67 & 77.89 \\
\midrule
GUI-CIDER-8B & 95.56 & 91.67 & 96.39 & 89.57 & 89.84 & 88.13 & 92.00 \\
\midrule
\rowcolor{gray!15} \multicolumn{8}{c}{\textbf{Hard Level}} \\
GPT-4o & 60.69 & 60.38 & 52.42 & 45.27 & 50.93 & 50.83 & 53.49 \\
Qwen-Max-VL  & 66.64 & 67.59 & 65.80 & 60.23 & 58.78 & 65.34 & 63.69 \\
Qwen2.5-VL-72B & 70.68 & 68.91 & 70.98 & 57.59 & 53.94 & 68.10 & 64.56 \\
UI-TARS-72B-DPO & 31.48 & 35.87 & 24.19 & 36.33 & 58.13 & 19.94 & 35.78 \\
InternVL3-72B  & 75.08 & 77.44 & 76.19 & 70.37 & 75.73 & 78.11 & 75.70 \\
\midrule
GUI-CIDER-8B & 93.33 & 90.48 & 94.33 & 92.17 & 88.20 & 93.15 & 91.83 \\
\bottomrule
\end{tabular}
\caption{Results of MMBench-GUI L1 benchmark. GUI-CIDER enhances GUI understanding. Although the data sources do not cover every platform in MMBench-GUI, improvements are still observed across all of them.}
\label{tab:mmbench}
\end{table*}

\paragraph{Proof of Property 3.}
For the product \(d \cdot g(x)\):
\begin{align}
    \frac{\partial}{\partial d} \big( d \cdot g \big) 
    &= g + d \cdot \frac{\partial g}{\partial d} \nonumber \\
    &= \lambda f + \frac{1 - \lambda f}{1 + \alpha d} \;-\; \frac{\alpha d (1 - \lambda f)}{(1 + \alpha d)^2} \nonumber \\
    &= \lambda f + (1 - \lambda f) \frac{1 + \alpha d - \alpha d}{(1 + \alpha d)^2} \nonumber \\
    &= \lambda f + \frac{1 - \lambda f}{(1 + \alpha d)^2}.
\end{align}

Since \(\lambda f \ge 0\) and \(1 - \lambda f \ge 0\) with a strictly positive denominator, we obtain \(\frac{\partial}{\partial d}(d \cdot g) > 0\) for all valid parameter settings. This guarantees that if two samples have densities \(d_1 < d_2\), then \(d_1 \cdot g(x_1) < d_2 \cdot g(x_2)\) (all else being equal), faithfully preserving the original density ordering of the semantic space.

\paragraph{Proof of Property 4.} 
The cross-partial derivative of $g(x)$ is:

\begin{equation}
\frac{\partial^2 g}{\partial f \partial d} = \frac{\partial}{\partial d} \left( \lambda \frac{\alpha d}{1+\alpha d} \right) = \frac{\lambda \alpha}{(1+\alpha d)^2}.
\end{equation}

Given $\lambda > 0$ and $\alpha > 0$, we strictly have $\frac{\partial^2 g}{\partial f \partial d} > 0$. 
This guarantees that the marginal utility of causal saliency $f$ increases monotonically with density $d$, effectively prioritizing high-quality reasoning structures within redundant clusters.

\section{Experiment}

\begin{table*}[t]
\centering
\setlength{\tabcolsep}{2pt}
\begin{tabular}{lccccccccc}
\toprule
\multirow{2}{*}{\textbf{Model}} 
& \multicolumn{3}{c}{\textbf{Interface Knowledge}} 
& \multicolumn{3}{c}{\textbf{Interaction Knowledge}} 
& \multicolumn{2}{c}{\textbf{Procedure Knowledge}} 
& \multirow{2}{*}{\textbf{Overall}} \\
\cmidrule(lr){2-4} \cmidrule(lr){5-7} \cmidrule(lr){8-9}
& state & widget & layout 
& effect & type & parameter 
& objective & workflow 
&  \\
\midrule

O3 & 83.03 & 84.12 & 88.39 & 74.83 & 75.98 & 45.75 & 69.45 & 95.47 & 73.30 \\

Gemini-2.5-Pro & 81.19 & 84.36 & 87.10 & 71.03 & 73.25 & 46.97 & 67.72 & 92.56 & 71.69 \\

GPT-5-Chat & 78.90 & 84.12 & 88.39 & 71.55 & 71.55 & 43.85 & 68.98 & 91.26 & 70.97 \\


Claude-Sonnet-4.5 & 74.77 & 81.52 & 82.58 & 49.83 & 70.19 & 43.33 &70.30 & 91.56 & 66.53 \\

Qwen3-VL-8B-Instruct & 66.97 & 76.30 & 79.35 & 60.86 & 63.54 & 45.06 & 71.02 & 78.96 & 65.23 \\

Qwen2.5-VL-72B & 69.27 & 77.49 & 80.00 & 61.72 & 64.91 & 38.99 & 62.20 & 85.44 & 63.88 \\

Doubao-V-Pro & 72.48 & 83.65 & 81.29 & 67.24 & 75.64 & 41.07 & 33.07 & 94.17 & 63.42 \\

Claude-Sonnet-4 & 70.18 & 78.44 & 78.06 & 41.90 & 62.52 & 42.11 & 65.20 & 94.82 & 62.16 \\

Qwen2.5-VL-7B & 53.21 & 67.77 & 60.00 & 51.72 & 50.60 & 39.34 & 16.22 & 48.87 & 45.16 \\

UI-TARS-1.5-7B & 49.54 & 59.48 & 59.35 & 22.24 & 59.11 & 34.32 & 38.74 & 55.34 & 44.27 \\

GUI-owl-7B & 60.09 & 64.93 & 63.23 & 21.55 & 55.37 & 36.05 & 21.26 & 39.81 & 40.74 \\

GLM-4.5 & 49.54 & 48.10 & 53.55 & 27.07 & 17.55 & 35.53 & 28.98 & 91.91 & 38.10\\ 
\midrule
GUI-CIDER-8B & 72.61 & 75.83& 80.13& 61.55&65.42 &46.45 & 71.81& 80.58 & 66.51
\\
\bottomrule
\end{tabular}
\caption{Results of GUI knowledge benchmark. The performance of GUI-CIDER-8B is close to Claude-Sonnet-4.5.}
\label{tab:gui_knowledge_bench}
\end{table*}

In this section, we first introduce the implementation of our GUI-CIDER experiments, then present the main results and provide analysis.

\subsection{Implementation}

\paragraph{Dataset.}
As shown in Table~\ref{tab:dataset_overview}, we conduct extensive experiments on three benchmarks, AITZ~\cite{zhang2024android}, AndroidControl~\cite{li2024effects}, and GUI-Odyssey~\cite{lu2025guiodyssey}, for GUI agent task completion, where the agent is required to output actions to accomplish tasks, and two benchmarks, MMbench-GUI L1~\cite{wang2025mmbench} and GUI knowledge bench~\cite{shi2025gui}, for GUI agent knowledge, both of which adopt the formats of multiple-choice questions (MCQs) and true-false (T/F) questions.

\paragraph{Evaluation Method.}
We used GUI-CIDER for data synthesis on the AITZ, AndroidControl, and GUI-Odyssey datasets.
The base models were Qwen3-VL-4B-Instruct and Qwen3-VL-8B-Instruct~\cite{Qwen3-VL}. 
In the main results section, we refer to the models obtained by mid-training Qwen3-VL-4B-Instruct and Qwen3-VL-8B-Instruct with GUI-CIDER as GUI-CIDER-4B and GUI-CIDER-8B, respectively.
For evaluation on MMBench-GUI L1 and GUI Knowledge Bench, we performed mid-training using a mixture of all synthesized data.
For evaluation on AITZ, AndroidControl, and GUI-Odyssey, we conducted mid-training using the data synthesized from the corresponding dataset.
Meanwhile, we adopted SFT as the baseline method for post-training.

\paragraph{Metrics.}
For the AITZ, AndroidControl, and GUI-Odyssey datasets, we report action type accuracy (type), step-wise success rate (SR), and task success rate (TSR).
For MMbench-GUI L1 and GUI Knowledge Bench, we compute the accuracy of multiple-choice questions and true-false questions under different subsets.

\subsection{Main Results}

The results on the AITZ, AndroidControl, and GUI-Odyssey datasets are shown in Table~\ref{tab:threebench}, the results on MMbench-GUI L1 are shown in Table~\ref{tab:mmbench}, and the results on the GUI knowledge benchmark are shown in Table~\ref{tab:gui_knowledge_bench}.

Based on the above results, we find:

(i) As shown in Table~\ref{tab:threebench}, mid-training yields gains in task completion capability across models of different parameter scales. Furthermore, when post-training is applied after mid-training, the benefits of GUI-CIDER still manifest. 
In addition, a 4B-scale GUI agent, after undergoing mid-training and post-training with GUI-CIDER, surpasses its 8B-scale counterpart, suggesting that for GUI agents, what matters may not be parameter scaling but rather knowledge scaling.

(ii) As shown in Table~\ref{tab:mmbench}, GUI-CIDER-8B significantly outperforms the baselines, indicating that GUI-CIDER brings improvements to the GUI content understanding capability of GUI agents.

(iii) As shown in Table~\ref{tab:gui_knowledge_bench}, overall, GUI-CIDER-8B clearly bridges the knowledge gap in GUI tasks, achieving performance close to that of Claude-Sonnet-4.5 at the 8B scale (66.51 vs. 66.53). 
Moreover, GUI-CIDER-8B surpasses all larger-scale models (e.g., o3, Gemini-2.5-Pro) on the objective subset (which assesses whether a task is truly completed), demonstrating that GUI-CIDER equips the GUI agent with a better understanding of tasks.

\section{Further Analysis}
In this section, we first compare the differences between models that have undergone post-training in the GUI agent domain and general models when used as the base model for GUI-CIDER, followed by an ablation study.
\subsection{Model Comparison Analysis}

We conduct an analysis to verify whether a GUI-specialized model that has already been post-trained in the GUI agent domain can acquire new world knowledge again through mid-training. 
We perform experiments on the AITZ dataset with OS-Atlas-pro-7B following the GUI-CIDER, and report results with the amount of GUI-CIDER-generated data increasing in 20\% increments. 

\begin{figure}[t]
    \centering
    \includegraphics[width=\linewidth]{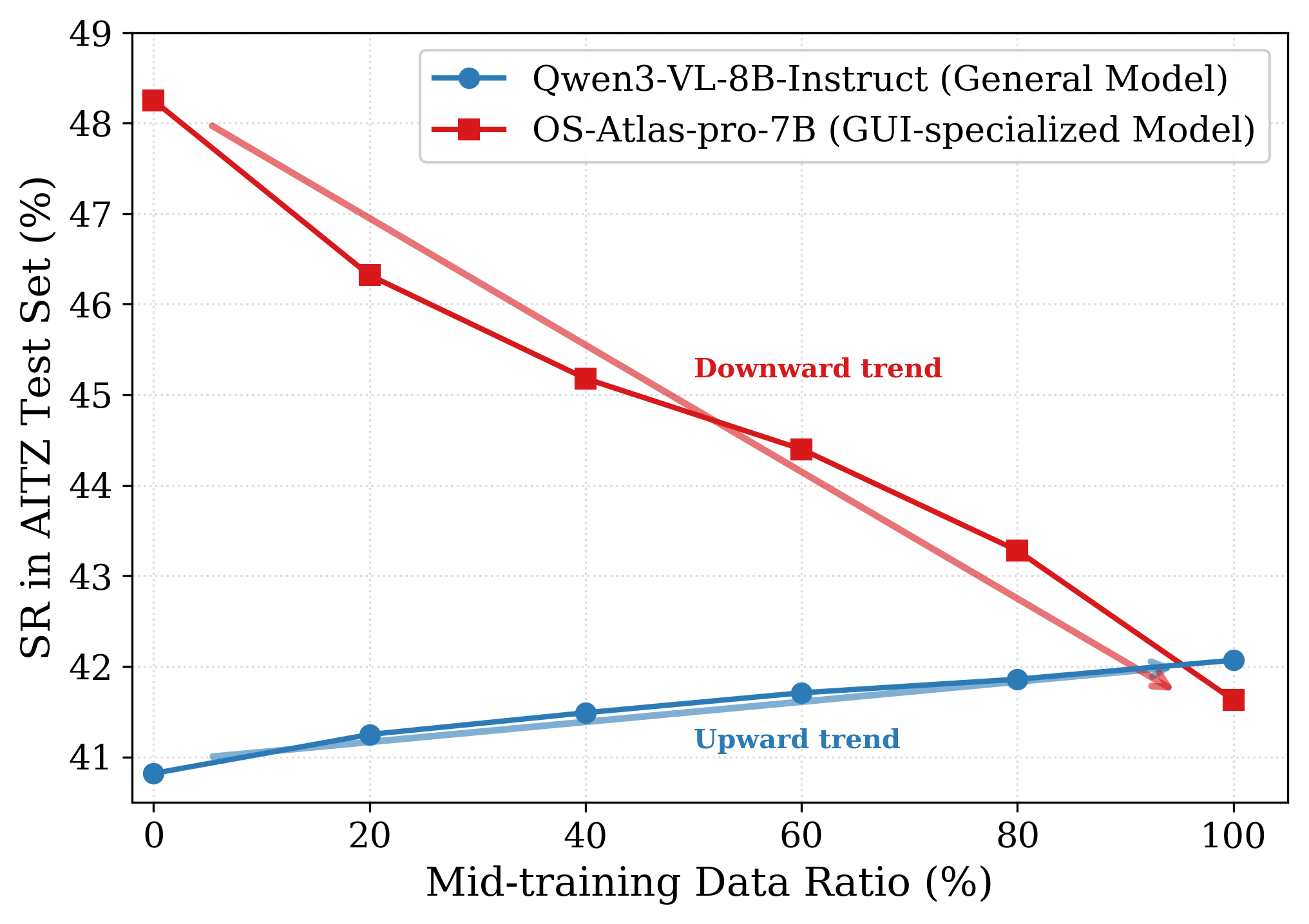}
    \caption{Comparison between a general model and a GUI-specialized model as the base model for mid-training, illustrating that models excessively post-trained in the GUI agent domain are no longer suitable for acquiring world knowledge through mid-training.}
    \label{model_comparison}
\end{figure}

As shown in Figure~\ref{model_comparison}, when using the general model Qwen3-VL-8B-Instruct as the base model, the GUI agent's SR consistently improves as more GUI-CIDER-generated data are incorporated. 
In contrast, when using OS-Atlas-pro-7B as the base model, the GUI agent's performance steadily declines. 
This is because OS-Atlas-pro-7B has undergone extensive post-training for GUI agents, which has already partially disrupted its original language representation capacity, making it difficult to learn new world knowledge through mid-training.
Therefore, performing mid-training on world knowledge before conducting post-training in the GUI agent domain would be a reasonable paradigm.

\subsection{Ablation Study}

\begin{table}[t]
\centering
\caption{Ablation study of exemplar reselection stage.}
\label{tab:ablation_study}
\setlength{\tabcolsep}{3pt}
\begin{tabular}{lcc}
\toprule
\textbf{Model} & \textbf{w/ Stage 2} & \textbf{w/o Stage 2} \\
\midrule
Qwen3-VL-4B-Instruct & 43.45 & 41.06 \\
Qwen3-VL-8B-Instruct & 48.55 & 42.34 \\
\bottomrule
\end{tabular}
\end{table}
We conduct an ablation study to investigate the necessity of the exemplar reselection stage in GUI-CIDER. 
Specifically, we compare SR after mid-training with the complete GUI-CIDER pipeline against a variant that removes the exemplar reselection stage on the GUI-Odyssey dataset.
As shown in Table~\ref{tab:ablation_study}, removing the exemplar reselection stage leads to a substantial drop in SR. 
This is because directly incorporating large-scale unscreened data into mid-training introduces a considerable amount of low-quality and redundant samples. 
Such noisy supervision can mislead the GUI agent and encourage shortcut or hacking behaviors, ultimately harming generalization and decision-making capability.


\section{Conclusion}
In this paper, we present \textsc{GUI-CIDER}, a mid-training framework that internalizes GUI world knowledge into GUI agents through causal internalization and density-aware exemplar reselection. 
Instead of relying on expensive external scaffolding or directly applying post-training to raw trajectories, GUI-CIDER synthesizes static planning knowledge and dynamic causal transition logic from GUI trajectories, then selects knowledge-rich and non-redundant exemplars with a retention function.
Experiments on three task completion benchmarks and two GUI knowledge benchmarks show that GUI-CIDER improves both GUI operation understanding and downstream task success. 
These results suggest that knowledge scaling is a promising path toward more capable GUI agents.
\section*{Limitations}
Due to computational resource constraints, GUI-CIDER employs LoRA instead of full parameter tuning during the mid-training stage. Additionally, the model parameters used for training range from 4B to 8B. Future work will further explore the effectiveness under full parameter tuning and scale the approach to larger models.


\bibliography{custom}

\appendix

\section{End-to-End Algorithm}
\label{sec:appendix_algorithm}

Algorithm~\ref{alg:guicider} summarizes the full GUI-CIDER pipeline, including knowledge-rich sample synthesis, density-aware exemplar reselection, and mid-training on the retained corpus.

\begin{algorithm}[t]
\small
\caption{GUI-CIDER}
\label{alg:guicider}
\begin{algorithmic}[1]
\Statex \textbf{Input:} $\mathcal{D}$, $\mathcal{M}_{exp}$, $\phi$, $K$, $\alpha$, $\lambda$, $\gamma$
\Statex \textbf{Output:} $\mathcal{X}_{\text{high}}$, $\theta$
\State Initialize synthesized corpus $\mathcal{X} \gets \emptyset$
\ForAll{$\tau=(T,\{(s_t,a_t,v_t)\}_{t=1}^{L}) \in \mathcal{D}$}
    \State $S \gets \mathcal{P}(T;\mathcal{M}_{exp})$
    \For{$t=1$ to $L$}
        \State $a_t^{nl} \gets \mathcal{B}(a_t, v_t)$
        \State Convert $s_{t-1}, s_t$ to text descriptions $d_{t-1}, d_t$
        \State $R_t \gets \mathcal{C}(T,d_{t-1},a_t^{nl},d_t \mid \mathcal{M}_{exp})$
        \State $x_t \gets \langle T, S, a_t^{nl}, R_t \rangle$
        \State $\mathcal{X} \gets \mathcal{X} \cup \{x_t\}$
    \EndFor
\EndFor
\ForAll{$x \in \mathcal{X}$}
    \State $f(x) \gets \tanh(\mathcal{K}(x)/\gamma)$
    \State Compute $r(x)$ by the ratio of local $K$-NN distance to global mean distance
\EndFor
\State Min-max normalize $\{r(x)\}_{x \in \mathcal{X}}$ into $\{d(x)\}_{x \in \mathcal{X}}$
\ForAll{$x \in \mathcal{X}$}
    \State $b_x \gets \frac{1}{1+\alpha d(x)}$
    \State $g(x) \gets b_x + \lambda f(x)(1-b_x)$
    \State Sample $\xi_x \sim \mathrm{Uniform}(0,1)$
    \If{$\xi_x \le g(x)$}
        \State $\mathcal{X}_{\text{high}} \gets \mathcal{X}_{\text{high}} \cup \{x\}$
    \EndIf
\EndFor
\State Mid-train agent parameters $\theta$ on $\mathcal{X}_{\text{high}}$ with the causal LM loss $\mathcal{L}_{\text{mid}}$
\State \Return $\mathcal{X}_{\text{high}}, \theta$
\end{algorithmic}
\end{algorithm}

\section{Additional Mathematical Proofs}
\label{sec:appendix_theory}

In this appendix, we provide supplementary theoretical results for the density-aware exemplar reselection rule used in GUI-CIDER. Throughout, let
\begin{equation}
    b(d) = \frac{1}{1+\alpha d},
\end{equation}
so that the retention function in Section~4 can be rewritten as
\begin{equation}
    \begin{aligned}
        g(x) &= b(d(x)) + \lambda f(x)\bigl(1-b(d(x))\bigr) \\
             &= \frac{1+\alpha \lambda f(x)d(x)}{1+\alpha d(x)}.
    \end{aligned}
\end{equation}
This form makes explicit that $g(x)$ interpolates between a density-based baseline $b(d)$ and a causal-saliency correction controlled by $\lambda$.

\subsection{Range, Boundary Cases, and Parameter Interpretation}

\begin{thmbox}
\textbf{Proposition A.1 (Range and Boundary Cases).}
For any sample $x$ with $f(x)\in[0,1]$, $d(x)\in[0,1]$, $\alpha>0$, and $\lambda\in[0,1]$, the retention score satisfies
\begin{equation}
    \frac{1}{1+\alpha d(x)} \le g(x) \le 1
\end{equation}
and
\begin{equation}
    \lambda f(x) \le g(x) \le 1.
\end{equation}
Moreover, the following boundary cases hold:
\begin{equation}
    g(x)=1 \quad \text{if } d(x)=0,
\end{equation}
\begin{equation}
    g(x)=\frac{1}{1+\alpha d(x)} \quad \text{if } f(x)=0,
\end{equation}
and
\begin{equation}
    g(x)=1 \quad \text{if } \lambda=1 \text{ and } f(x)=1.
\end{equation}
\end{thmbox}

\paragraph{Proof.}
Since $f(x)\in[0,1]$ and $\lambda\in[0,1]$, we have
\begin{equation}
    \begin{aligned}
        g(x) &= b(d(x))+\lambda f(x)\bigl(1-b(d(x))\bigr) \\
             &\ge b(d(x)) = \frac{1}{1+\alpha d(x)}.
    \end{aligned}
\end{equation}
Likewise,
\begin{equation}
    g(x)\le b(d(x))+\bigl(1-b(d(x))\bigr)=1.
\end{equation}
For the second lower bound,
\begin{equation}
    \begin{aligned}
        g(x)-\lambda f(x)
        &= b(d(x))\bigl(1-\lambda f(x)\bigr) \\
        &\ge 0
    \end{aligned}
\end{equation}
which implies $g(x)\ge \lambda f(x)$. The boundary identities follow directly by substitution into the closed form of $g(x)$.

\subsection{Proofs of the Four Desiderata}

\begin{thmbox}
\textbf{Proposition A.2 (Causal Monotonicity).}
For fixed $d(x)$, the retention function is monotonically non-decreasing in $f(x)$.
\end{thmbox}

\paragraph{Proof.}
For a fixed density $d$, the derivative of $g$ with respect to $f$ is
\begin{equation}
    \frac{\partial g}{\partial f}
    = \lambda \left(1-\frac{1}{1+\alpha d}\right)
    = \lambda \frac{\alpha d}{1+\alpha d}.
\end{equation}
Since $\lambda\ge 0$, $\alpha>0$, and $d\ge 0$, we have $\frac{\partial g}{\partial f}\ge 0$. Therefore, $g(x)$ is monotonically non-decreasing in $f(x)$. It is strictly increasing whenever $\lambda>0$ and $d>0$.

\begin{thmbox}
\textbf{Proposition A.3 (Density Penalty).}
For fixed $f(x)$, the retention function is monotonically non-increasing in $d(x)$.
\end{thmbox}

\paragraph{Proof.}
Differentiating $g$ with respect to $d$ gives
\begin{equation}
    \frac{\partial g}{\partial d}
    = -\frac{\alpha(1-\lambda f)}{(1+\alpha d)^2}.
\end{equation}
Because $f\in[0,1]$ and $\lambda\in[0,1]$, we have $1-\lambda f\ge 0$. Hence $\frac{\partial g}{\partial d}\le 0$, so denser samples always receive no larger retention scores.

\begin{thmbox}
\textbf{Proposition A.4 (Density Order Preservation).}
The reweighted density score $d \cdot g(x)$ is strictly increasing in $d$:
\begin{equation}
    \frac{\partial}{\partial d}\bigl(d \cdot g(x)\bigr) > 0.
\end{equation}
\end{thmbox}

\paragraph{Proof.}
Using the closed form of $g$,
\begin{equation}
    d\cdot g(x) = \frac{d+\alpha \lambda f d^2}{1+\alpha d}.
\end{equation}
Applying the quotient rule yields
\begin{equation}
    \begin{aligned}
        \frac{\partial}{\partial d}\bigl(d\cdot g(x)\bigr)
        &=
        \frac{(1+2\alpha \lambda f d)(1+\alpha d)}{(1+\alpha d)^2} \\
        &\quad -
        \frac{\alpha(d+\alpha \lambda f d^2)}{(1+\alpha d)^2}.
    \end{aligned}
\end{equation}
After simplification,
\begin{equation}
    \frac{\partial}{\partial d}\bigl(d\cdot g(x)\bigr)
    = \frac{1+2\alpha \lambda f d+\alpha^2 \lambda f d^2}{(1+\alpha d)^2}.
\end{equation}
Every term in the numerator is non-negative, and the constant term is strictly positive. Therefore,
\begin{equation}
    \frac{\partial}{\partial d}\bigl(d\cdot g(x)\bigr) > 0.
\end{equation}
Thus, the reweighted density preserves the original ordering induced by $d$.

\begin{thmbox}
\textbf{Proposition A.5 (Density-Causal Synergy).}
The retention rule exhibits positive interaction between causal saliency and density:
\begin{equation}
    \frac{\partial^2 g(x)}{\partial f \partial d} > 0.
\end{equation}
\end{thmbox}

\paragraph{Proof.}
From Proposition A.2,
\begin{equation}
    \frac{\partial g}{\partial f}
    = \lambda \frac{\alpha d}{1+\alpha d}.
\end{equation}
Differentiating again with respect to $d$ gives
\begin{equation}
    \begin{aligned}
        \frac{\partial^2 g}{\partial f \partial d}
        &= \lambda \alpha \frac{(1+\alpha d)-\alpha d}{(1+\alpha d)^2} \\
        &= \frac{\lambda \alpha}{(1+\alpha d)^2}.
    \end{aligned}
\end{equation}
For $\lambda>0$ and $\alpha>0$, the above quantity is strictly positive. Hence the marginal value of causal saliency becomes larger in denser regions, which is exactly the desired synergy effect.

\subsection{Expected Retained Corpus Size and Information Preservation}

The next result formalizes the intuition that stochastic retention preserves a non-trivial fraction of the original reasoning signal.

\begin{thmbox}
\textbf{Proposition A.6 (Expectation of the Retained Corpus).}
Let $\mathbf{1}_x$ be the Bernoulli indicator of whether sample $x$ is retained, namely $\mathbf{1}_x = 1$ if $\xi_x \le g(x)$ and $\mathbf{1}_x = 0$ otherwise. Then
\begin{equation}
    \mathrm{E}\bigl[|\mathcal{X}_{\text{high}}|\bigr]
    = \sum_{x\in\mathcal{X}} g(x).
\end{equation}
In particular,
\begin{equation}
    \sum_{x\in\mathcal{X}} \frac{1}{1+\alpha d(x)}
    \le
    \mathrm{E}\bigl[|\mathcal{X}_{\text{high}}|\bigr]
    \le
    |\mathcal{X}|.
\end{equation}
\end{thmbox}

\paragraph{Proof.}
By construction, $\mathbf{1}_x$ is a Bernoulli random variable with mean $g(x)$. Therefore,
\begin{equation}
    \begin{aligned}
        |\mathcal{X}_{\text{high}}|
        &= \sum_{x\in\mathcal{X}} \mathbf{1}_x, \\
        \mathrm{E}\bigl[|\mathcal{X}_{\text{high}}|\bigr]
        &= \sum_{x\in\mathcal{X}} \mathrm{E}[\mathbf{1}_x]
         = \sum_{x\in\mathcal{X}} g(x).
    \end{aligned}
\end{equation}
The bounds follow immediately from Proposition A.1.

\begin{thmbox}
\textbf{Proposition A.7 (Lower Bound on Preserved Causal Information).}
Define the causal information mass of a subset $\mathcal{A}\subseteq \mathcal{X}$ as
\begin{equation}
    I_f(\mathcal{A}) = \sum_{x\in\mathcal{A}} f(x).
\end{equation}
Then the expected preserved causal information ratio satisfies
\begin{equation}
    \begin{aligned}
        \frac{\mathrm{E}[I_f(\mathcal{X}_{\text{high}})]}{I_f(\mathcal{X})}
        &=
        \frac{\sum_{x\in\mathcal{X}} f(x)g(x)}{\sum_{x\in\mathcal{X}} f(x)} \\
        &\ge
        \lambda \frac{\sum_{x\in\mathcal{X}} f(x)^2}{\sum_{x\in\mathcal{X}} f(x)} \\
        &\ge \lambda \bar{f}
    \end{aligned}
\end{equation}
where $\bar{f} = \frac{1}{|\mathcal{X}|}\sum_{x\in\mathcal{X}} f(x)$ is the average causal saliency of the original corpus.
\end{thmbox}

\paragraph{Proof.}
Since $I_f(\mathcal{X}_{\text{high}})=\sum_x \mathbf{1}_x f(x)$, linearity of expectation gives
\begin{equation}
    \begin{aligned}
        \mathrm{E}[I_f(\mathcal{X}_{\text{high}})]
        &= \sum_{x\in\mathcal{X}} f(x)\mathrm{E}[\mathbf{1}_x] \\
        &= \sum_{x\in\mathcal{X}} f(x)g(x).
    \end{aligned}
\end{equation}
By Proposition A.1, $g(x)\ge \lambda f(x)$ for every $x$, hence
\begin{equation}
    \sum_{x\in\mathcal{X}} f(x)g(x)
    \ge
    \lambda \sum_{x\in\mathcal{X}} f(x)^2.
\end{equation}
Dividing both sides by $\sum_x f(x)$ proves the first inequality. For the second inequality, Cauchy--Schwarz implies
\begin{equation}
    \sum_{x\in\mathcal{X}} f(x)^2
    \ge
    \frac{\bigl(\sum_{x\in\mathcal{X}} f(x)\bigr)^2}{|\mathcal{X}|},
\end{equation}
and therefore
\begin{equation}
    \begin{aligned}
        \lambda \frac{\sum_x f(x)^2}{\sum_x f(x)}
        &\ge
        \lambda \frac{\sum_x f(x)}{|\mathcal{X}|} \\
        &= \lambda \bar{f}.
    \end{aligned}
\end{equation}
This shows that the thinning process retains a guaranteed fraction of the original causal signal in expectation.

\begin{figure*}[t]
    \centering
    \includegraphics[width=\linewidth]{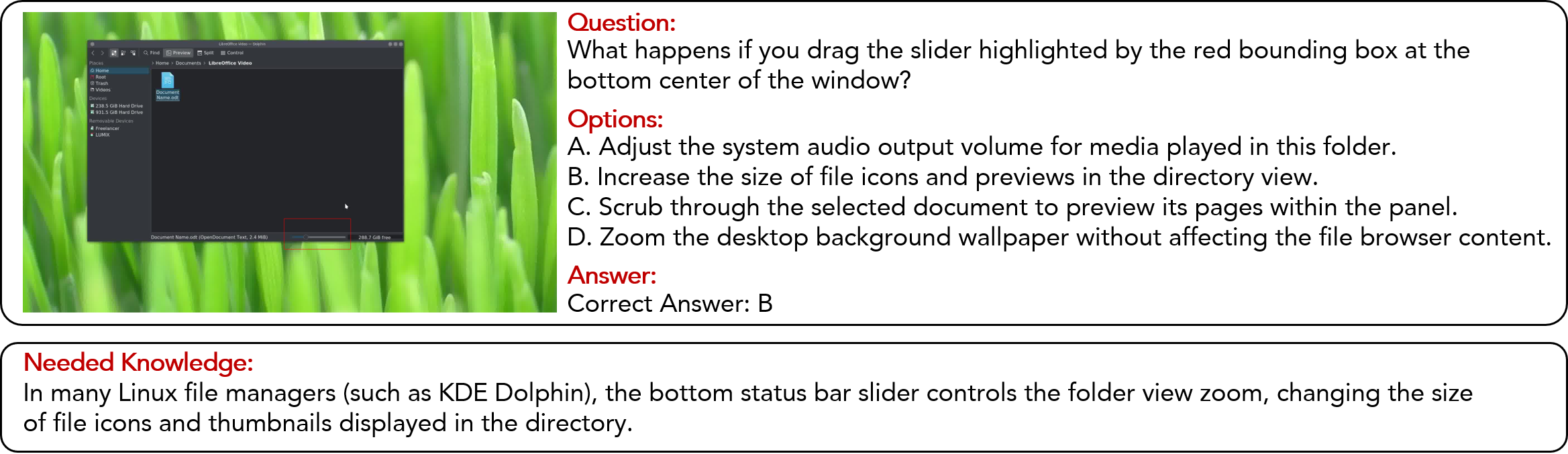}
    \caption{Examples from GUI Knowledge Bench.}
    \label{fig:guiknowledgebench}
\end{figure*}
\begin{figure*}[t]
    \centering
    \includegraphics[width=\linewidth]{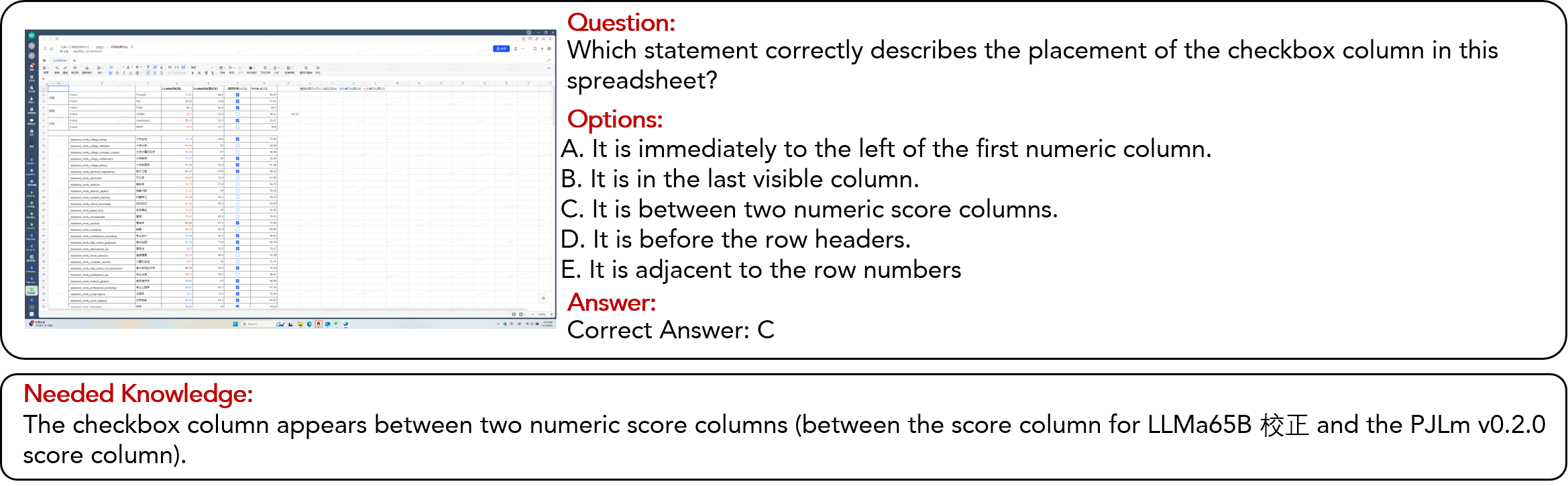}
    \caption{Examples from MMBench-GUI L1.}
    \label{fig:mmbenchgui}
\end{figure*}

\subsection{Stability Under Score Estimation Errors}

In practice, both $f(x)$ and $d(x)$ are estimated from synthesized text and embedding geometry. The next result shows that moderate perturbations in these quantities induce controlled perturbations in $g(x)$.

\begin{thmbox}
\textbf{Proposition A.8 (Lipschitz Stability).}
Let $\hat{f}(x)$ and $\hat{d}(x)$ be perturbed estimates satisfying
\begin{equation}
    |\hat{f}(x)-f(x)| \le \varepsilon_f,
    \qquad
    |\hat{d}(x)-d(x)| \le \varepsilon_d.
\end{equation}
If $\hat{g}(x)=g(\hat{f}(x),\hat{d}(x))$, then
\begin{equation}
    |\hat{g}(x)-g(x)|
    \le
    \alpha \varepsilon_d
    + \lambda \frac{\alpha}{1+\alpha}\varepsilon_f.
\end{equation}
\end{thmbox}

\paragraph{Proof.}
By the mean value theorem applied to the bivariate function $g(f,d)$, there exists a point on the line segment joining $(f,d)$ and $(\hat{f},\hat{d})$ such that
\begin{equation}
    \begin{aligned}
        |\hat{g}-g|
        &\le
        \sup |\partial_d g| \cdot |\hat{d}-d| \\
        &\quad +
        \sup |\partial_f g| \cdot |\hat{f}-f|.
    \end{aligned}
\end{equation}
From the derivatives established above,
\begin{equation}
    |\partial_d g|
    =
    \frac{\alpha(1-\lambda f)}{(1+\alpha d)^2}
    \le \alpha
\end{equation}
and
\begin{equation}
    \begin{aligned}
        |\partial_f g|
        &= \lambda \frac{\alpha d}{1+\alpha d} \\
        &\le \lambda \frac{\alpha}{1+\alpha}
    \end{aligned}
\end{equation}
because $d\in[0,1]$. Substituting the error bounds yields
\begin{equation}
    \begin{aligned}
        |\hat{g}(x)-g(x)|
        &\le \alpha \varepsilon_d \\
        &\quad + \lambda \frac{\alpha}{1+\alpha}\varepsilon_f.
    \end{aligned}
\end{equation}
Hence the retention rule is stable under bounded score estimation noise.

\begin{figure*}
    \centering
    \includegraphics[width=0.8\linewidth]{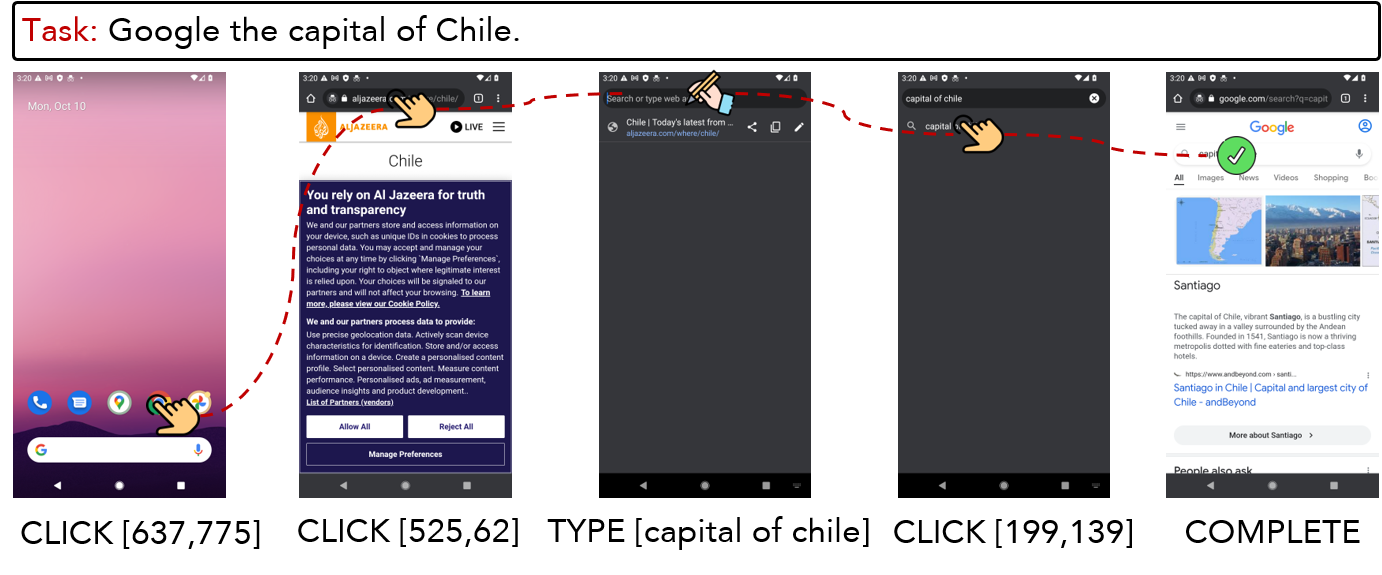}
    \caption{Examples from GUI agent task completion benchmarks.}
    \label{fig:aitz}
\end{figure*}
\begin{figure*}[t]
    \centering
    \includegraphics[width=\linewidth]{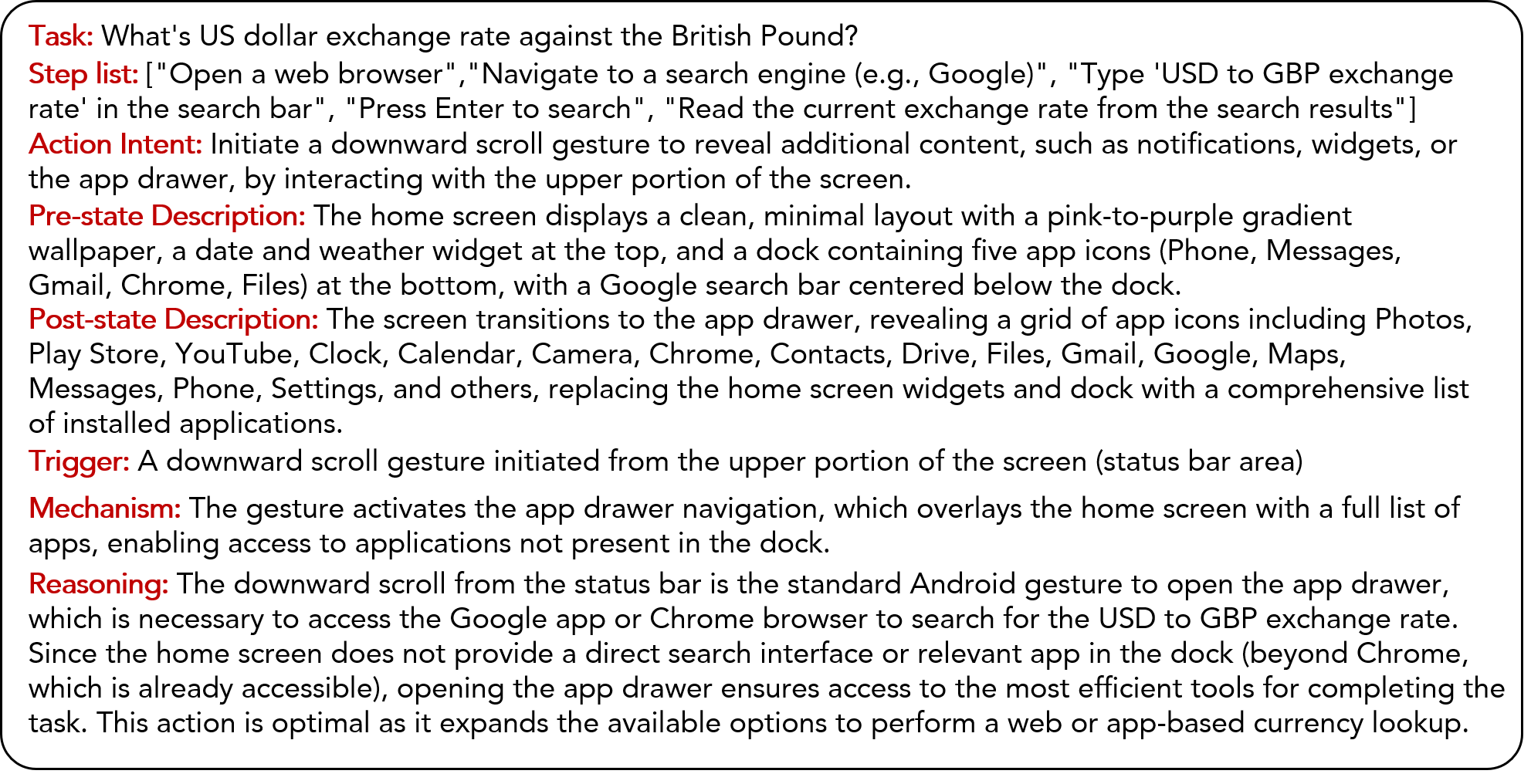}
    \caption{Examples of GUI-CIDER Synthetic Data.}
    \label{fig:case}
\end{figure*}

\section{Benchmark Details and Examples}
\noindent We briefly summarize the five benchmarks used in our experiments and point readers to the corresponding examples in the appendix.

\paragraph{GUI Knowledge Bench.}
GUI Knowledge Bench is a diagnostic benchmark for GUI-specific knowledge rather than end-to-end execution. It evaluates whether a model can understand widget functions, interface states, interaction effects, and workflow progress across diverse GUI platforms. Representative examples are shown in Figure~\ref{fig:guiknowledgebench}.

\paragraph{MMBench-GUI L1.}
MMBench-GUI is a hierarchical cross-platform benchmark for GUI agents, and the L1 split used here focuses on GUI content understanding. Its examples test whether a model can read interface content and reason about the semantics and relative placement of GUI elements. Representative examples are shown in Figure~\ref{fig:mmbenchgui}.

\paragraph{AITZ.}
AITZ (Android-In-The-Zoo) is an Android GUI navigation benchmark built from screen-action pairs with Chain-of-Action-Thought annotations. It evaluates whether an agent can infer the next GUI action from the current screen, prior context, and task instruction. A representative example is shown in Figure~\ref{fig:aitz}.

\paragraph{AndroidControl.}
AndroidControl studies real-world Android control at scale using human demonstrations paired with both high-level and low-level instructions. It is designed to evaluate how well agents follow everyday mobile tasks under realistic data diversity and task complexity.

\paragraph{GUI-Odyssey.}
GUI-Odyssey focuses on long-horizon cross-app mobile navigation, where successful completion requires carrying context across multiple apps and steps. Compared with single-app benchmarks, it places heavier demands on history tracking, planning, and cross-app reasoning. 

\section{Examples of Synthetic Data}

Figure~\ref{fig:case} presents a representative synthetic training example produced by GUI-CIDER. 
Starting from a task and a concrete GUI transition, GUI-CIDER organizes the annotation into a step list, action intent, pre-state description, post-state description, trigger, mechanism, and reasoning. 
This format converts raw interaction traces into structured supervision that captures not only what action should be taken, but also why the action is appropriate in the current GUI context.

\section{Keyword and Prompt Templates}
\label{sec:appendix_artifacts}

\begin{figure}[t]
    \centering
    \includegraphics[width=\linewidth]{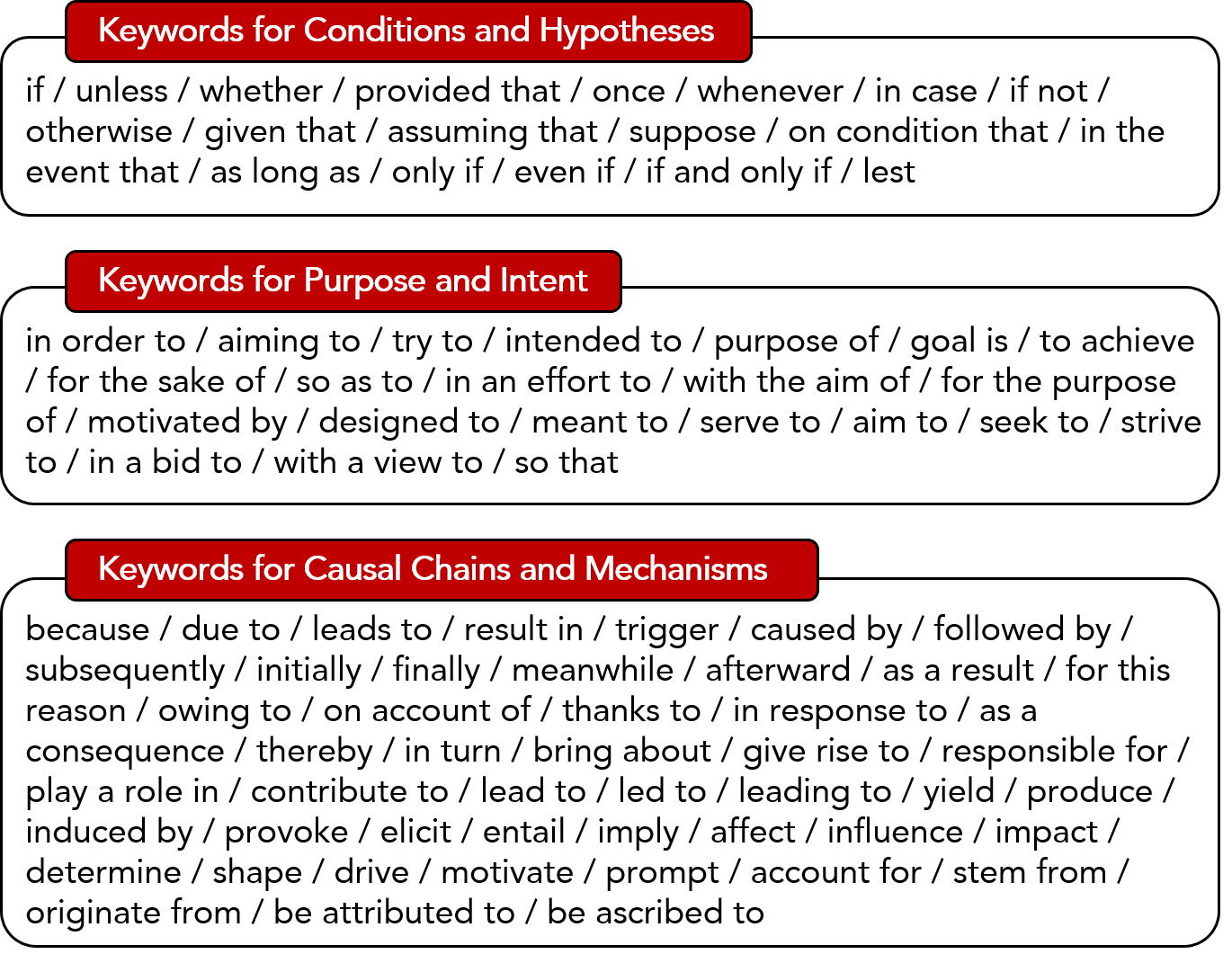}
    \caption{Keyword lexicon used in Stage 1 data synthesis (Part I). The listed markers target conditional and hypothetical reasoning, purpose and intent, and explicit causal chains or UI mechanisms. We use these lexical cues as lightweight anchors when eliciting and auditing reasoning-rich rationales from synthesized trajectories.}
    \label{fig:keywords_part1}
\end{figure}

\begin{figure}[t]
    \centering
    \includegraphics[width=\linewidth]{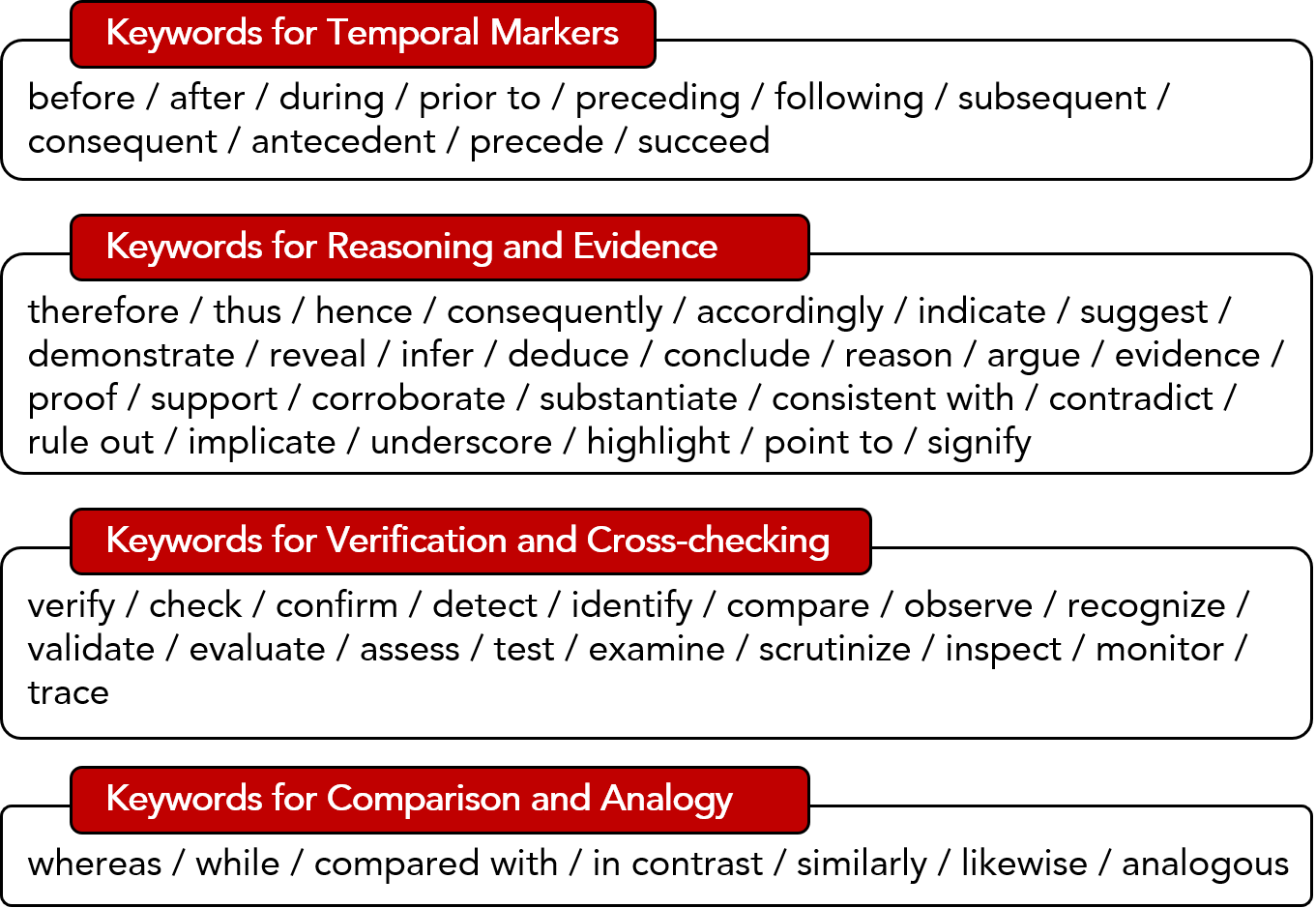}
    \caption{Keyword lexicon used in Stage 1 data synthesis (Part II). This part focuses on temporal markers, reasoning and evidence expressions, verification cues, and comparison or analogy terms, which help identify ordering constraints, supporting evidence, and contrastive reasoning in textualized GUI transitions.}
    \label{fig:keywords_part2}
\end{figure}

Figures~\ref{fig:keywords_part1} and~\ref{fig:keywords_part2} summarize the lexical categories used in our data-synthesis pipeline to surface reasoning-rich textual patterns. These categories cover conditional and hypothetical statements, purpose and intent, explicit causal chains, temporal ordering, evidential language, verification cues, and comparison markers. In our pipeline, they are used as lightweight anchors for prompt design and quality inspection, helping the expert model generate rationales that more consistently expose triggers, mechanisms, and outcome-oriented reasoning.

\begin{figure}[t]
    \centering
    \includegraphics[width=\linewidth]{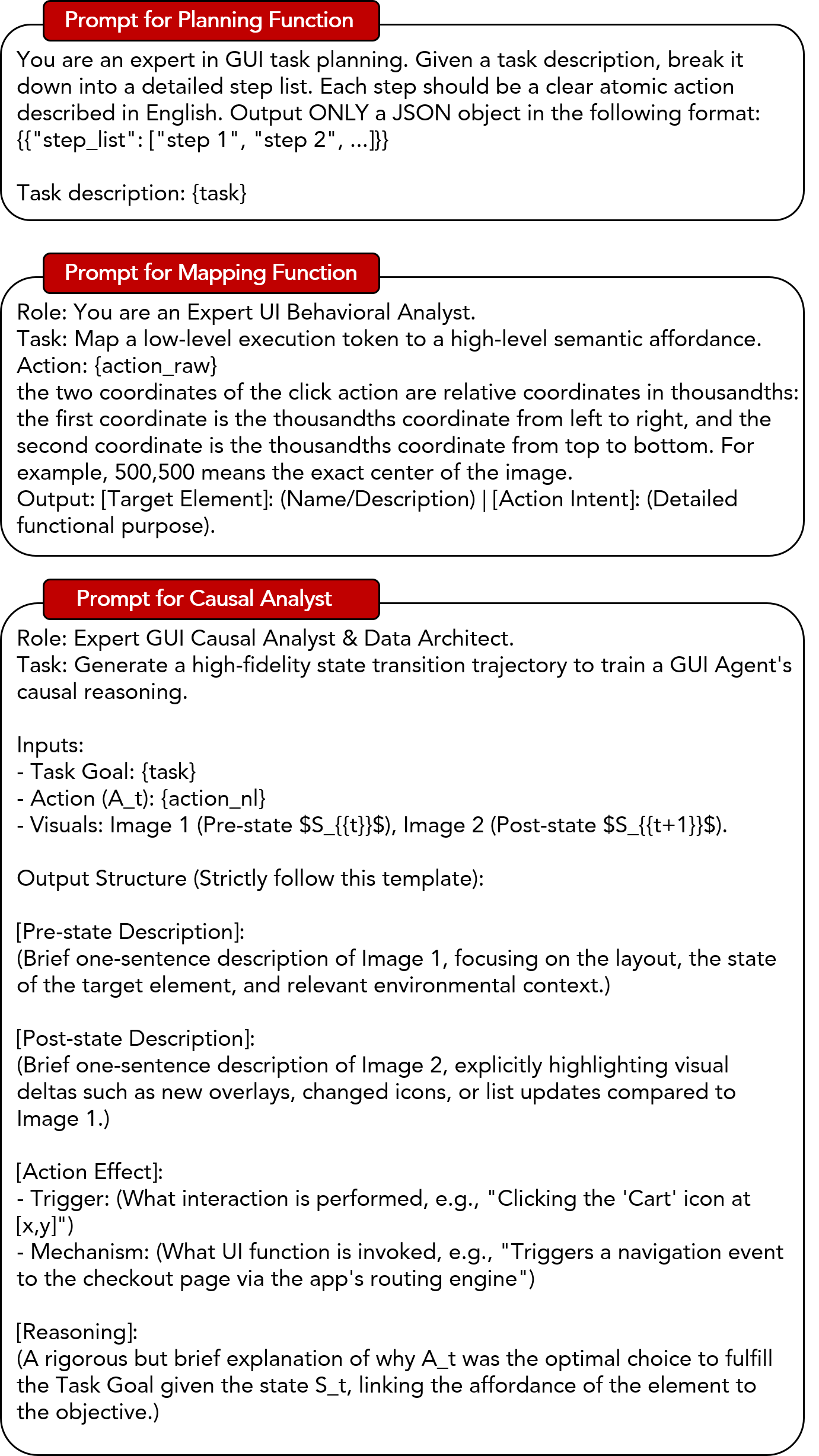}
    \caption{Prompt templates used to instantiate the planning function $\mathcal{P}$, the semantic mapping function $\mathcal{B}$, and the causal analyst $\mathcal{C}$.}
    \label{fig:prompt}
\end{figure}

Figure~\ref{fig:prompt} presents the prompt templates used to instantiate the three key modules in Stage 1: the planning function $\mathcal{P}$, the semantic behavioral grounding function $\mathcal{B}$, and the causal analyst $\mathcal{C}$. Together, these templates standardize how raw task descriptions, low-level GUI actions, and paired pre-/post-state screenshots are converted into the textual tuple $\langle T, S, a_t^{nl}, R_t \rangle$ used in mid-training.

\begin{table}[t]
    \centering
    \small
    \begin{tabular}{c l}
     \toprule
    \textbf{Action Type} & \textbf{Action Description} \\ 
    \midrule 
    CLICK & Click at specified position. \\
    TYPE & Enter text at designated location. \\
    SCROLL & Scroll in direction. \\
    PRESS\_BACK & Go to previous screen. \\
    PRESS\_HOME & Go to home page. \\
    ENTER & Press enter button. \\
    OPEN\_APP & Open specified app. \\
    WAIT & Wait for screen to load. \\
    LONG\_PRESS & Long press at specified position. \\
    COMPLETE & Indicate task finished. \\
    IMPOSSIBLE & Indicate task impossible. \\
    \bottomrule
    \end{tabular}
    \caption{Action space in our experiment.}
    \label{action_space}
\end{table}

\section{Experimental Details}
In our experiments, the Planning function is implemented using deepseek-v4-flash, while both the mapping function and the causal analyst are based on Qwen3-VL-32B-Instruct. 
The base models are Qwen3-VL-4B-Instruct and Qwen3-VL-8B-Instruct. 
We set the per-device training batch size to 4, gradient accumulation steps to 2, learning rate to 1.0e-5, and train for 2 epochs. 
Moreover, for the post-training mentioned in the main text, we used a full-scale SFT approach, using the same training set as GUI-CIDER for training and the same test set for evaluation.
For the same data, applying mid-training with GUI-CIDER followed by post-training yields improvements compared to direct post-training.
This demonstrates that GUI-CIDER can further unlock the potential of the data.
The total computational cost amounts to 1,400 hours on 80G GPUs.
For GUI agent task completion benchmarks, we adhere to the assessment methods of existing works: for actions with coordinates such as CLICK and LONG\_PRESS, a error of less than 14\% is considered correct. For TYPE actions, an F1 score greater than 0.5 is required to be counted as correct. In all other cases, exact matching is necessary for correctness.
And TSR for a task will be 1 only if SR for every single frame within that task is 1.
Action space is shown as Table~\ref{action_space}.

\end{document}